%% file: main.tex
\documentclass[conference]{IEEEtran}
\IEEEoverridecommandlockouts

\usepackage{cite}
\usepackage{amsmath,amssymb,amsfonts}
\usepackage{algorithmic}
\usepackage{graphicx}
\usepackage{textcomp}
\usepackage{xcolor}
\usepackage{booktabs}
\usepackage{multirow}
\usepackage{makecell}

\def\BibTeX{{\rm B\kern-.05em{\sc i\kern-.025em b}\kern-.08em
    T\kern-.1667em\lower.7ex\hbox{E}\kern-.125emX}}
\begin{document}

\title{SAM-OCTA2: Layer Sequence OCTA Segmentation with Fine-tuned Segment Anything Model 2}

\input{sections/a_authors}

\maketitle

\input{sections/0_abstract}
\input{sections/1_introduction}
\input{sections/2_related_work}
\input{sections/3_method}
\input{sections/4_experiments}
\input{sections/5_conclusion}

\section*{Acknowledgment}

This work is supported by the Chongqing Natural Science Foundation Innovation and Development Joint Fund (CSTB2023NSCQ-LZX0109), Chongqing Technology Innovation \& Application Development Key Project (cstb2022tiad-kpx0148), and Fundamental Research Funds for the Central Universities (No.2022CDJYGRH-001).

\bibliographystyle{IEEEtran}
\bibliography{icassp2025.bib}

\end{document}

%% file: sections/a_authors.tex

\author{\IEEEauthorblockN{1\textsuperscript{st} Xinrun Chen}
\IEEEauthorblockA{\textit{College of Computer Science} \\
\textit{Chongqing University}\\
Chongqing, China \\
chenxinrun@cqu.edu.cn}
\and
\IEEEauthorblockN{2\textsuperscript{nd} Chengliang Wang*}
\IEEEauthorblockA{\textit{College of Computer Science} \\
\textit{Chongqing University}\\
Chongqing, China \\
wangcl@cqu.edu.cn}
\and
\IEEEauthorblockN{3\textsuperscript{rd} Haojian Ning}
\IEEEauthorblockA{\textit{College of Computer Science} \\
\textit{Chongqing University}\\
Chongqing, China \\
nhj@cqu.edu.cn}
\and
\IEEEauthorblockN{4\textsuperscript{th} Mengzhan Zhang}
\IEEEauthorblockA{\textit{Department of Ophthalmology} \\
\textit{Xiang’an Hospital of Xiamen University}\\
Xiamen, China \\
2459851117@qq.com}
\and
\IEEEauthorblockN{5\textsuperscript{th} Mei Shen}
\IEEEauthorblockA{\textit{Department of Ophthalmology} \\
\textit{Xiang’an Hospital of Xiamen University}\\
Xiamen, China \\
luckymay2889@163.com}
\and
\IEEEauthorblockN{6\textsuperscript{th} Shiying Li}
\IEEEauthorblockA{\textit{Department of Ophthalmology} \\
\textit{Xiang’an Hospital of Xiamen University}\\
Xiamen, China \\
shiying\_li@126.com}
}


%% file: sections/0_abstract.tex
**Notice:** This work has been submitted to the IEEE for possible publication. Copyright may be transferred without notice, after which this version may no longer be accessible.

\begin{abstract}
Segmentation of indicated targets aids in the precise analysis of optical coherence tomography angiography (OCTA) samples. Existing segmentation methods typically perform on 2D projection targets, making it challenging to capture the variance of segmented objects through the 3D volume. To address this limitation, the low-rank adaptation technique is adopted to fine-tune the Segment Anything Model (SAM) version 2, enabling the tracking and segmentation of specified objects across the OCTA scanning layer sequence. To further this work, a prompt point generation strategy in frame sequence and a sparse annotation method to acquire retinal vessel (RV) layer masks are proposed. This method is named SAM-OCTA2 and has been experimented on the OCTA-500 dataset. It achieves state-of-the-art performance in segmenting the foveal avascular zone (FAZ) on regular 2D en-face and effectively tracks local vessels across scanning layer sequences. The code is available at: https://github.com/ShellRedia/SAM-OCTA2.

\end{abstract}

\begin{IEEEkeywords}
OCTA, image segmentation, fine-tuning, segment anything model, sparse annotation.
\end{IEEEkeywords}

%% file: sections/1_introduction.tex
\section{Introduction}
\label{Sec_Intro}

OCTA is a crucial technology for visualizing the retinal vascular system, particularly the microvascular structures and blood flow dynamics \cite{javed2023optical}. It provides detailed, non-invasive imaging of retinal structures and has been widely applied to analyze and diagnose retinal diseases such as age-related macular degeneration, branch retinal vein occlusion, diabetic retinopathy, and glaucoma \cite{taylor2024role, xue2024soul, nouri2024optical, braun2024role}. OCTA captures high-resolution volumetric samples by stacking B-scans for depth, while en-face projections are created by slicing the volume across layers \cite{meiburger2021automatic}.

Segmenting RVs and FAZ in OCTA is crucial for assessing retinal health and diagnosing diseases. Extensive deep learning-based segmentation methods have been developed and have demonstrated strong performance. Existing methods can be classified into 2D and 3D types based on the input format. The 2D methods take single or several slice projected images, with advantages in processing efficiency and lightweight design \cite{sharma2022lightweight, hu2022joint, tan2023oct2former}. The 3D methods use full volumetric as input, performing better segmentation but demanding higher computational resources such as time and memory \cite{li2020ipn, wu2021paenet, zhong2022dive, quan2025multi}. However, constrained by annotation, both types of methods currently predict targets on en-face or B-scan projections.

SAM is the most powerful foundational zero-shot segmentation model for addressing natural image tasks \cite{kirillov2023segment}. With retraining or fine-tuning methods, SAM has been applied in medical images with impressive performance \cite{ma2024segment, wu2023medical}. SAM 2 is an extended version of SAM for video segmentation tasks \cite{ravi2024sam}. With prompts on any frame of a video to specify a target of interest, it enables segmenting of the target throughout the entire frame sequence. The SAM-OCTA effectively segmented local vessels on en-face OCTA images with fine-tuned SAM, demonstrating the feasibility of utilizing SAM 2 on OCTA data\cite{wang2024sam}.

We find that the layer scanning structure of OCTA samples corresponds well to the frame sequence input of SAM 2. Inspired by this, we call our method SAM-OCTA2 and summarize the contributions as follows:

\begin{enumerate}
    \item Applying low-rank adaptation (LoRA) technology for SAM 2 fine-tuning enables it to perform effective local RV or FAZ segmentation across layer sequences.
    \item A corresponding prompt point generation strategy is proposed to identify and indicate a local object.
    \item A sparse annotation method is designed to provide layer RV annotations for the OCTA volume samples.
\end{enumerate}

%% file: sections/2_related_work.tex
\section{Related Work}

\subsection{OCTA Segmentation Models}

Most OCTA segmentation models have adopted custom-designed modules and processing strategies to accommodate the distribution and shape of the biomarkers, especially RVs. The attention mechanism and transformer layers are well-suited for RV segmentation due to the ability to capture long-range dependencies and global connectivity, which is essential for accurately modeling the complex branching structures of RVs \cite{vaswani2017attention}. For the capacity to handle varying shapes and sparse distributions, methods such as OCTA-Net, FARGO, and ARP-Net et al. introduce the attention modules to achieve precise segmentation of both large and fine vessels across the retina \cite{ma2020rose, peng2021fargo, quan2025multi, liu2023transformer, jiang2024octa, tan2023oct2former, zhu2022ovs}. Some more methods make efforts on data balancing, parameter reduction, and detail preservation with developed techniques achieving promising segmentation results on OCTA datasets \cite{ma2022retinal, ning2024accurate, wang2023db, li2022image}. These methods show that the OCTA deep networks widely adopt the modified transformer layers and achieve accurate segmentation of RV and FAZ.

\subsection{SAM 2 and Parameter-Efficient Fine-tuning Techniques}

SAM 2, as a foundational segmentation model, has been pre-trained on over 50K video samples. Its zero-shot feature allows easy transfer to various applications through limited prompts. While SAM2 excels in semantic understanding of regular frame sequences, fine-tuning is essential to adapt it for OCTA feature extraction. An ideal fine-tuning method should achieve two goals: improving OCTA segmentation performance and maintaining the previous module cooperation. Therefore, parameter-efficient fine-tuning techniques such as inserting adapter layers or using LoRA are feasible options \cite{pfeiffer2020adapterfusion, hu2021lora}.

%% file: sections/3_method.tex
\section{Method}

\begin{figure}
  \centering
  \includegraphics[width=1\linewidth]{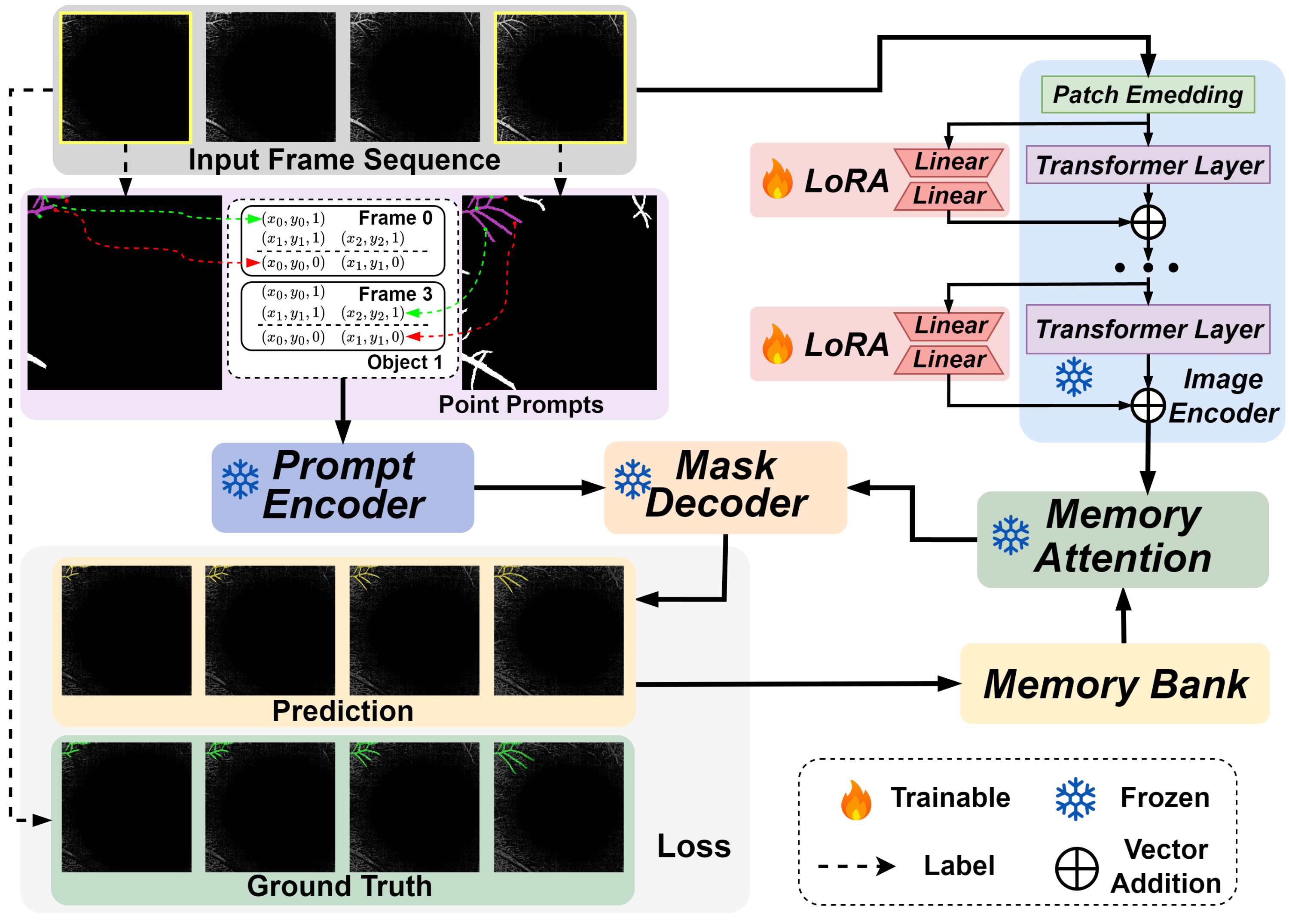}
  \caption{The schematic diagram of the SAM-OCTA2 structure. The original model weights are frozen to preserve the semantic understanding and image processing capabilities through pre-training. The memory bank is essentially a queue and does not contain trainable parameters.}
  \label{Fig_Architecture}
\end{figure}

In this paper, we proposed the SAM-OCTA2 by fine-tuning the pre-trained SAM 2 with the OCTA dataset. This model performs flexible OCTA segmentation in both en-face projection and layer sequence images, and the fine-tuning process is shown in Fig. \ref{Fig_Architecture}. The SAM is composed of an image encoder, a flexible prompt encoder, and a fast mask decoder to support the prompt conditional input. Two additional modules, namely memory bank and memory attention, are introduced in SAM 2 to integrate information from multiple frames. 

\subsection{Fine-tuning of SAM 2}
\label{sec_fine_tune_sam2}

The image encoder extracts the semantics of input frames with stacked transformer layers, which is well-suited for OCTA images. The prompt encoder encodes the input prompts (points, boxes, masks) into conditional vectors to indicate the segmentation target in the image sequence. In this work, only the point prompts are utilized for simplicity. The mask decoder maps the embeddings of the image sequence, prompt, and memorized features to a segmentation mask. The output mask is used for loss calculation and passed to the memory bank for multi-frame feature fusion. The memory bank uses a FIFO queue storing several produced frames from the mask decoder to retain past predictions and prompt information. The memory attention module fuses the features of the current frame and the past features stored in the memory bank by stacked transformer blocks. It fuses features by calculating self-attention for each frame and cross-attention between different frames.

The proportion of trainable parameters in each module of SAM 2 with the base configuration quantified as follows: image encoder: 85.703\%, prompt encoder: 0.007\%, mask decoder: 5.227\%, and memory attention: 9.063\%. Only the image encoder is fine-tuned with LoRA since it contains most of the parameters \cite{hu2021lora}. All the trainable parameters of the original SAM 2 are frozen first, and the LoRA module's blocks are added as side branches to the transformer layers of the image encoder. The blocks of the LoRA module are lightweight linear layers, which account for 1.68\% of the total parameters of the entire model, and only the LoRA parameters are updated during fine-tuning.

\subsection{Prompt Points Generation Strategy}

\begin{figure}
  \centering
  \includegraphics[width=1\linewidth]{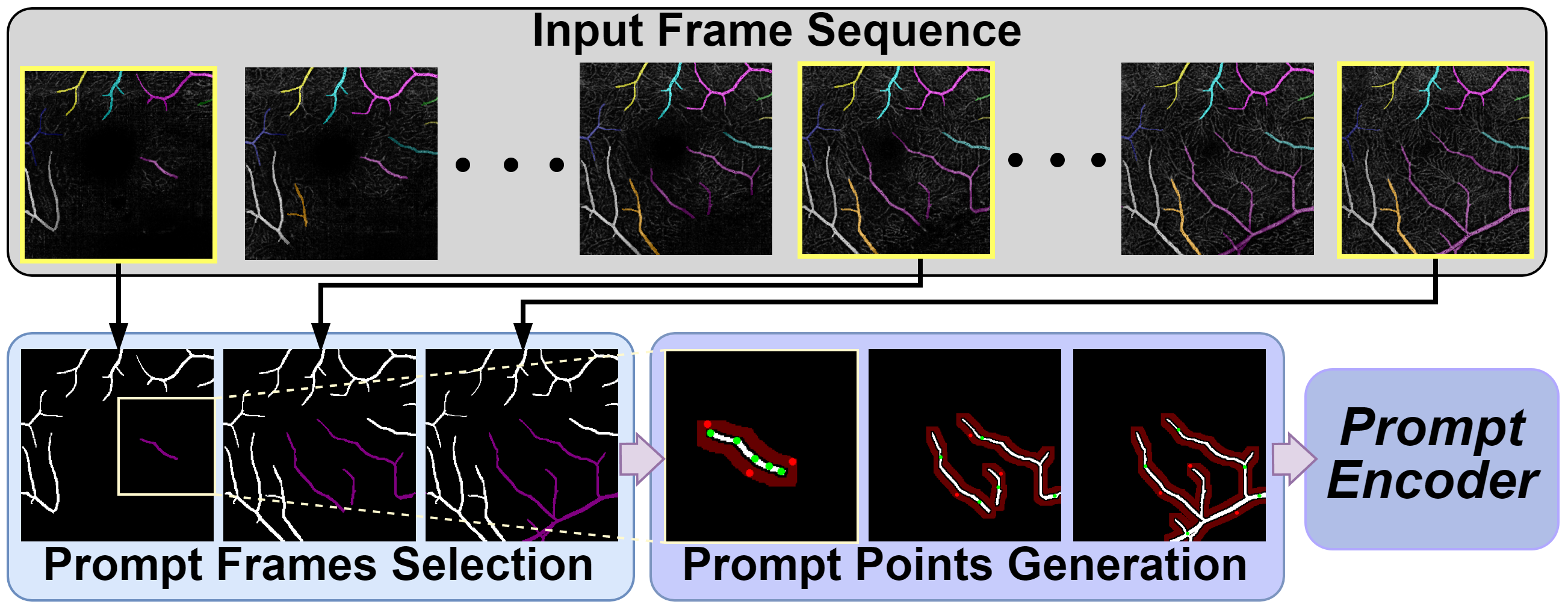}
  \caption{The illustration of prompt point generation in the scanning layers of OCTA samples. Each vessel is represented by a distinct color. The purple vessel is selected as the segmentation target. The red regions surrounding vessels are designated to propose negative points.}
  \label{Fig_PromptGeneration}
\end{figure}

The prompt points of SAM 2 include four elements: frame, object, type, and coordinate. These elements describe how a prompt point tracks a specified object within the image sequence. The process of generating prompt points for OCTA samples is shown in Fig \ref{Fig_PromptGeneration}. We first select one or several frames and find an object that appears in all the selected frames as the segmented target. The coordinates of the prompt point depend on its type. If the prompt point is positive, the coordinate is sampled within the target pixel. If negative, the coordinate is chosen from the target's surrounding region, which is calculated using the dilation operation. Additionally, a separation gap of three pixels width is set between the positive and negative regions to reduce ambiguity.

In this work, RV and FAZ are segmented in en-face OCTA images from sequential scanning layers, and each layer corresponds to a frame in the image sequence. Identifying the same object across different layers is essential. The FAZ is unique to a sample and does not require any additional processing. For RV segmentation, each visible vessel or vascular cluster is independently distinguished. The thickness and position of the same vessel across multiple layers are nearly consistent, with only the visible length varying. Utilizing this property, each vessel can be labeled using the calculation of connected components based on the en-face projection RV annotation. Since the segmentation of scanning layers does not follow anatomical structures, an object might be dispersed into multiple connected components. Each connected component contains at least one prompt point in the generation process, if possible.

\subsection{Layer Annotation of Retinal Vessel}

\begin{figure}
  \centering
  \includegraphics[width=1\linewidth]{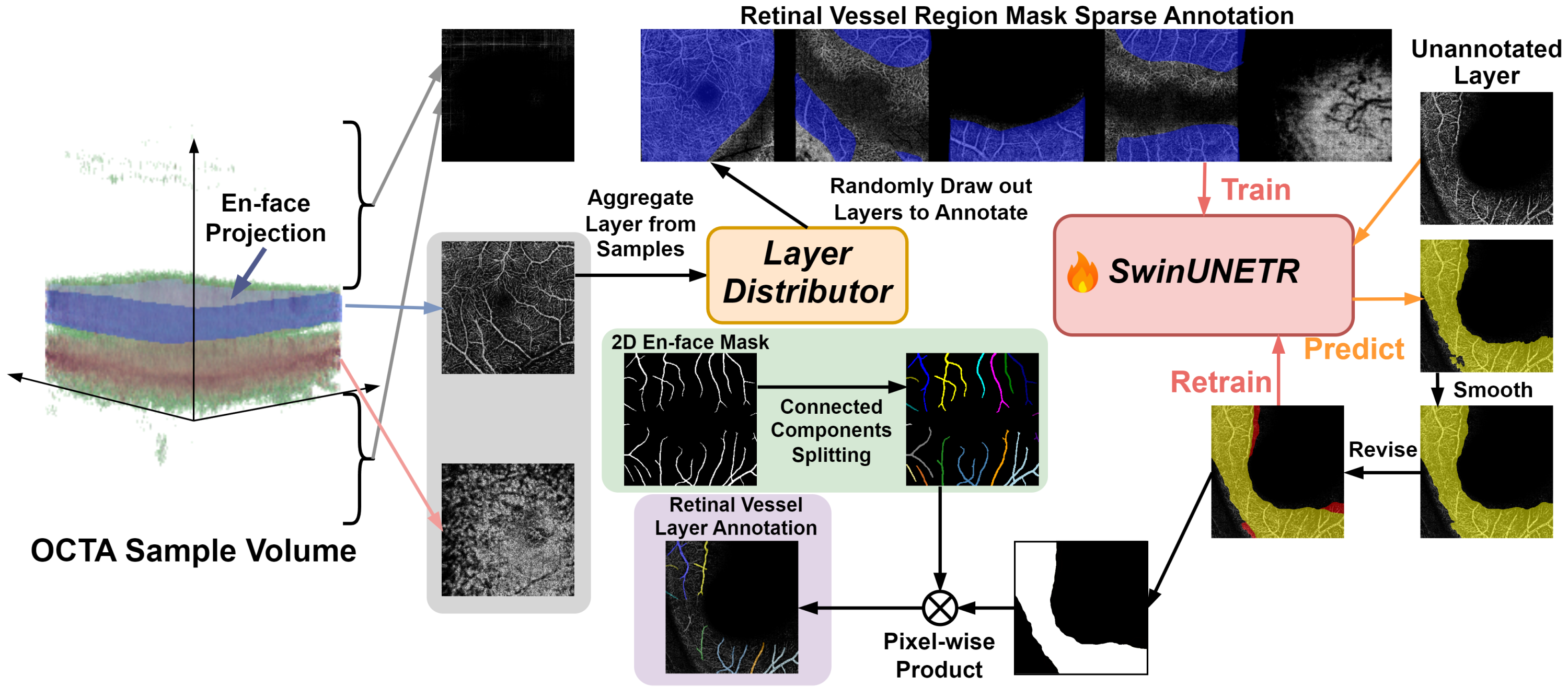}
  \caption{Illustration of sparse annotation for retinal vessel in scanning layers. We segment the regions where vessels appear instead of segmenting the RVs directly. This strategy utilizes existing annotations to enhance accuracy. The layer distributor randomly selects batches of potential RV scanning layers for manual annotation. For regions, blue represents the manually annotated, yellow indicates the model-predicted, and red denotes the modified. The predicted region is smoothed by the Gaussian filter.}
  \label{Fig_SparseAnnotation}
\end{figure}

Current public OCTA datasets lack layer segmentation annotations for RV, so we designed a sparse annotation method to address this gap, as illustrated in Fig. \ref{Fig_SparseAnnotation}. In an OCTA volume sample, most scanning layers are either blank or missing vessels, so we screened and discarded the blank layers. Then, we aggregated all the reserved layers and randomly sampled 1,000 layers for manual annotation of vessel regions with masks. The annotated layers were used to train the SwinUNETR segmentation model implemented by the MONAI library \cite{hatamizadeh2021swin, cardoso2022monai}. The predicted results were manually inspected, and layers with obvious errors were revised and added to the training set for model retraining. This process was repeated multiple times until the segmentation results were sufficiently accurate. The final layer RV annotation was obtained by performing the intersection operation between masks of en-face RV and the predicted region of each layer.

%% file: sections/4_experiments.tex
\section{Experiments}

\subsection{Dataset and Settings}

The dataset used in this paper is OCTA-500 \cite{li2024octa}. It is the largest publicly available OCTA dataset and the only one that provides 3D scanning layers. This dataset contains 500 OCTA samples in 3D format and 2D en-face projection layers. It offers FAZ but lacks RV in 3D annotation and provides complete 2D annotation for RV, FAZ, capillary, artery, and vein. The samples are divided into two subsets based on the field of view (FoV): $3mm \times 3mm$ (3M) and $6mm \times 6mm$ (6M), containing 200 and 300 samples, respectively. The data augmentation strategies include horizontal flipping and random slight rotation implemented by the Albumentations tool \cite{info11020125}.

Our SAM-OCTA2 is deployed on an A100 graphic card with 80 GB memory. The optimizer used is AdamW, and the learning rate is $5 \times 10^{-6}$. The loss function is Dice loss. The division of the training and test sets follows the IPN-v2's configuration \cite{li2020ipn} for comparison. For en-face projection image segmentation, the results were compared with previous work, while for layer sequence segmentation, only ablation studies were conducted due to the lack of existing related research. In the sequence training stage, the input frames are sampled at equal intervals from the scanning layers of the same OCTA sample, and the frame length ranges from 4 to 8. From the sampled frames, 1 to 3 frames are selected to generate prompt points, with the priority order as the first frame, the last frame, and the middle frame. Only one object is marked with prompt points in each segmentation, with 1 to 10 positive points and 0 to 6 negative points. The evaluation metrics are averaged across the segmentation results of all objects in the frame sequence.

\subsection{Results}

The segmentation results using metrics Dice and Jaccard, which are calculated as follows:

\begin{equation} \label{Eq_Dice_1}
Dice(\hat{Y}, Y) = \frac{2 |\hat{Y} \cap Y|}{|\hat{Y}| + |Y|},
\end{equation}
\begin{equation}\label{Eq_Jaccard_2}
Jaccard(\hat{Y}, Y) = \frac{|\hat{Y} \cap Y|}{|\hat{Y} \cup Y|}.
\end{equation}

where $Y, \hat{Y}$ → the ground-truth and predicted value.

\begin{figure}
  \centering
  \includegraphics[width=1\linewidth]{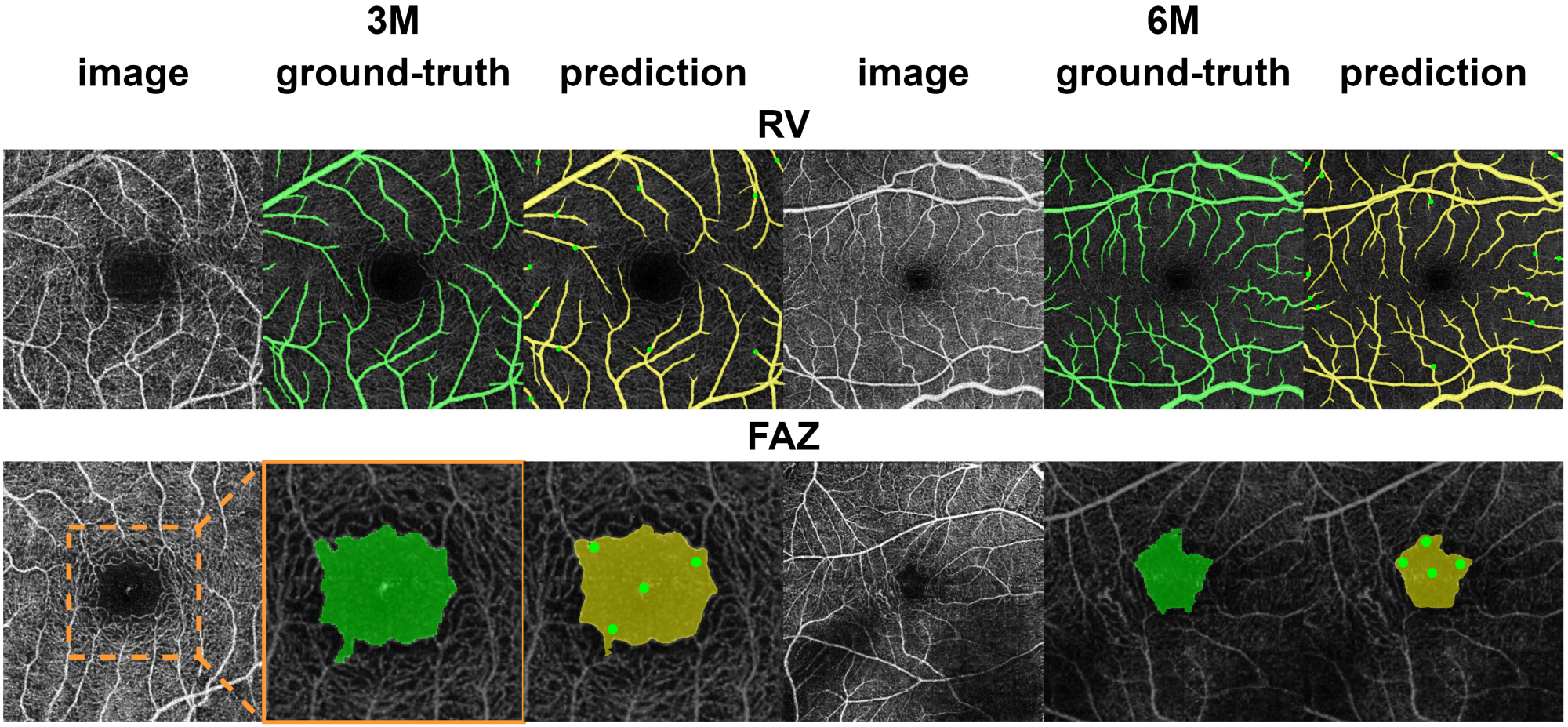}
  \caption{Segmentation samples of RV and FAZ on en-face OCTA images. The FAZ region has been enlarged for clearer observation, and the same applies to Fig.\ref{Fig_ResultSamples_Sequence} below.}
  \label{Fig_ResultSamples_Projection}
\end{figure}

RV and FAZ segmentation on en-face projected labels are regular tasks in previous studies, and we summarize the comparative results in Table \ref{Table_Segmentation2D}. The cited works have undergone detailed experiments and are more pertinent to this study \cite{li2020ipn, peng2021fargo, hu2022joint, wang2024sam}. The visualized results are presented in Fig. \ref{Fig_ResultSamples_Projection}. Our method achieves precise segmentation of targets on the en-face projection images and approaches state-of-the-art comprehensive performance.

\input{tables/Segmentation2D_result}

 For the layer sequence segmentation, we selected four types of conditions in quantity: frame length, prompted frames, and positive and negative points, with values of 4, 2, 5, and 3 in the baseline setting. In the ablation study, each condition was individually modified, and the results are shown in Table \ref{Table_SequenceAblation}. 

The prompt point input on partial frames can basically achieve target localization and segmentation across the entire layer sequence. Similar to the results of the en-face projection task, it is easier to segment on the 3M subset layer sequence segmentation. However, the impact of FoVs on target types is the opposite in these two tasks. The layer scanning more readily splits RVs into multiple parts, resulting in decreased segmentation performance. The splitting ruins the segmentation details, such as boundary and connectivity. As the input prompt information increases, including both prompt frames and prompt points, the segmentation performance typically improves. An unexpected result is that increasing the input frame length improves FAZ segmentation, even without additional prompt information.

\input{tables/SequenceAblation_result}

\begin{figure}
  \centering
  \includegraphics[width=1\linewidth]{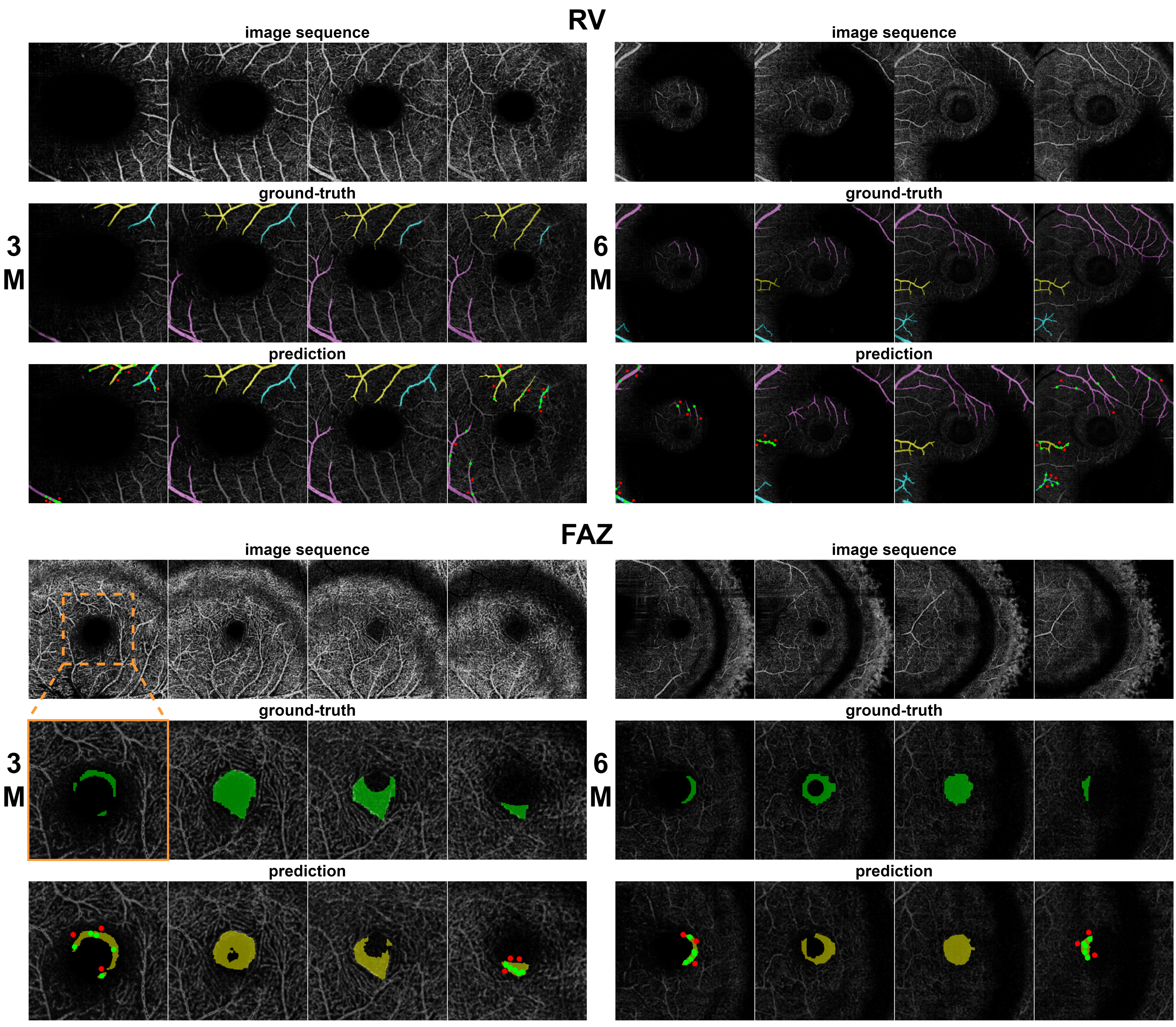}
  \caption{Samples of layer sequence segmentation with four frames. For simplicity, only three vessels are shown in the RV segmentation, distinguished by different colors. Note that each vessel is predicted separately, and the figure merges the results for visualization.}
  \label{Fig_ResultSamples_Sequence}
\end{figure}

%% file: tables/Segmentation2D_result.tex
\begin{table}[t]
\centering
\caption{RV and FAZ en-face segmentation results on OCTA-500 dataset(underscores indicate the top two highest values).}
\label{Table_Segmentation2D}
\begin{tabular}{cccccc}
\toprule
\multicolumn{2}{c}{Label} & \multicolumn{2}{c}{RV} & \multicolumn{2}{c}{FAZ} \\
Method & Metric & 3M & 6M & 3M & 6M \\
\midrule
    U-Net &    Dice ↑ & 0.9068 & 0.8876 & 0.9747 & 0.8770 \\
   (2015) & Jaccard ↑ & 0.8301 & 0.7987 & 0.9585 & 0.8124 \\
\cmidrule(lr){2-6}
  IPN V2+ &    Dice ↑ & \underline{0.9274} & \underline{0.8941} & 0.9755 & 0.9084 \\
   (2020) & Jaccard ↑ & \underline{0.8667} & \underline{0.8095} & 0.9532 & 0.8423 \\
\cmidrule(lr){2-6}
    FARGO &    Dice ↑ & 0.9168 & 0.8915 & \underline{0.9839} & \underline{0.9272} \\
   (2021) & Jaccard ↑ & 0.8470 & 0.8050 & 0.9684 & \underline{0.8701} \\
\cmidrule(lr){2-6}
Joint-Seg &    Dice ↑ & 0.9113 & \underline{0.8972} & \underline{0.9843} & 0.9051 \\
   (2022) & Jaccard ↑ & 0.8378 & \underline{0.8117} & \underline{0.9693} & 0.8473 \\
\cmidrule(lr){2-6}
 SAM-OCTA &    Dice ↑ & 0.9199 & 0.8869 & 0.9838 & 0.9073 \\
   (2024) & Jaccard ↑ & \underline{0.8520} & 0.7975 & \underline{0.9692} & 0.8473 \\
\midrule
SAM-OCTA2 &    Dice ↑ & \underline{0.9207} & 0.8923 & 0.9833 & \underline{0.9284} \\
   (ours) & Jaccard ↑ & 0.8428 & 0.8046 & 0.9687 & \underline{0.8733} \\
\bottomrule
\end{tabular}
\end{table}

%% file: tables/SequenceAblation_result.tex
\begin{table}[t]
\centering
\caption{Layer sequence segmentation results on OCTA-500 dataset under diverse input conditions}
\label{Table_SequenceAblation}
\begin{tabular}{ccccccc}
\toprule
\multicolumn{3}{c}{Label} & \multicolumn{2}{c}{RV} & \multicolumn{2}{c}{FAZ} \\
\multicolumn{2}{c}{Condition} & Metric & 3M & 6M & 3M & 6M \\
\midrule
\multicolumn{2}{c}{\multirow{2}{*}{Baseline}} & Dice ↑ & 0.6833 & 0.5487 & 0.7001 & 0.6828 \\
&& Jaccard ↑ & 0.5667 & 0.4428 & 0.5653 & 0.5399  \\
\cmidrule(lr){1-7}
\multirow{4}{*}{\makecell{Frame \\ Length}} & \multirow{2}{*}{6} & Dice ↑ & 0.6965 & 0.5447 & 0.7333 &  0.7047 \\
&& Jaccard ↑ & 0.5719 & 0.4402 & 0.6069 & 0.5633 \\
\cmidrule(lr){3-7}
& \multirow{2}{*}{8} & Dice ↑ & 0.6960 & 0.5478 & 0.7412 & 0.7141 \\
&& Jaccard ↑ & 0.5705 & 0.4435 & 0.6156 & 0.5750 \\
\cmidrule(lr){2-7}
\multirow{4}{*}{\makecell{Prompt \\ Frames}} & \multirow{2}{*}{1} & Dice ↑ & 0.6611 & 0.5273 & 0.5958 & 0.5810 \\
&& Jaccard ↑ & 0.5277 & 0.4101 & 0.4789 & 0.4556 \\
\cmidrule(lr){3-7}
& \multirow{2}{*}{3} & Dice ↑ & 0.7088 & 0.5837 & 0.7315 & 0.7045 \\
&& Jaccard ↑ & 0.5710 & 0.4426 & 0.6021 & 0.5633 \\
\cmidrule(lr){2-7}
\multirow{4}{*}{\makecell{Positive \\ Points}} & \multirow{2}{*}{1} & Dice ↑ & 0.6518 & 0.5156 & 0.6714 & 0.6480 \\
&& Jaccard ↑ & 0.5165 & 0.4048 & 0.5371 & 0.5057 \\
\cmidrule(lr){3-7}
& \multirow{2}{*}{10} & Dice ↑ & 0.6871 & 0.5544 & 0.7124 & 0.6934 \\
&& Jaccard ↑ & 0.5506 & 0.4278 & 0.5792 & 0.5503 \\
\cmidrule(lr){2-7}
\multirow{4}{*}{\makecell{Negative \\ Points}} & \multirow{2}{*}{0} & Dice ↑ & 0.6730 & 0.5404 & 0.6924 & 0.6689 \\
&& Jaccard ↑ & 0.5359 & 0.4152 & 0.5567 & 0.5262 \\
\cmidrule(lr){3-7}
& \multirow{2}{*}{6} & Dice ↑ & 0.6851 & 0.5510 & 0.7112 & 0.6844 \\
&& Jaccard ↑ & 0.5484 & 0.4248 & 0.5783 & 0.5406 \\
\bottomrule
\end{tabular}
\end{table}

%% file: sections/5_conclusion.tex
\section{Conclusion}

We propose a method called SAM-OCTA2 for both layer sequence and projection segmentation in an OCTA volume or a single image. With minimal prompt input, SAM-OCTA2 enables tracking local targets in OCTA data within 2D or volume space. We believe this is a flexible and highly promising method that helps in optical disease diagnosis and 3D structure reconstruction of samples.